\documentclass{Odyssey2026}

\interspeechcameraready

\title{MEGConformer: Conformer-Based MEG Decoder for Robust Speech and Phoneme Classification}

\author[affiliation={1}]{Xabier}{de Zuazo}
\author[affiliation={1}]{Ibon}{Saratxaga}
\author[affiliation={1}]{Eva}{Navas}

\affiliation{Communications Engineering}
            {HiTZ Center, University of the Basque Country -- UPV/EHU}
            {Spain}
\email{xabier.dezuazo@ehu.eus, ibon.saratxaga@ehu.eus, eva.navas@ehu.eus}

\usepackage{comment}
\usepackage{soul}
\begin{document}

\maketitle

\begin{abstract}

    Decoding speech-related information from non-invasive MEG is a key step toward scalable brain-computer interfaces. We present compact Conformer-based decoders on the LibriBrain 2025 PNPL benchmark for two core tasks: Speech Detection and Phoneme Classification. Our approach adapts a compact Conformer to raw 306-channel MEG signals, with a lightweight convolutional projection layer and task-specific heads. For Speech Detection, a MEG-oriented SpecAugment provided a first exploration of MEG-specific augmentation. For Phoneme Classification, we used inverse-square-root class weighting and a dynamic grouping loader to handle 100-sample averaged examples. In addition, a simple instance-level normalization proved critical to mitigate distribution shifts on the holdout split. Using the official Standard track splits and F1-macro for model selection, our best systems achieved 88.9\% (Speech) and 65.8\% (Phoneme) on the leaderboard, winning the Phoneme Classification Standard track.
\end{abstract}

\section{Introduction}

In the emerging fields of deep learning and neuroscience, decoding speech from non-invasive brain signals remains a central problem in neural signal processing and brain-computer interfaces (BCIs)~\cite{panachakel2021,dehgan2025,murad2025}. The recently released LibriBrain dataset~\cite{ozdogan2025} and its associated PNPL 2025 Competition~\cite{landau2025} provide a unique opportunity to address this challenge by enabling large-scale within-subject modeling of magnetoencephalography (MEG) recordings. LibriBrain contains over 50 hours of continuous MEG data from a single participant listening to the complete Sherlock Holmes corpus, accompanied by precise voice activity and phoneme-level alignments. The competition defines two foundational decoding tasks: Speech Detection, which distinguishes between speech and silence, and Phoneme Classification, which predicts the phoneme being perceived from averaged MEG signals. Both tasks include a Standard and an Extended track, where the Standard track restricts participants to using only the official training data. We participated exclusively in the Standard track, and our system won the Phoneme Classification task in the final competition results.

Our approach explores an initial adaptation of state-of-the-art Automatic Speech Recognition (ASR) architectures to non-invasive neural signals. Specifically, we leverage the Conformer model~\cite{gulati2020}, a convolution-augmented Transformer~\cite{vaswani2017} that combines the global context modeling of self-attention~\cite{lin2017} with the local feature extraction strength of convolutional networks~\cite{lecun2015}. Conformer-based architectures have been and continue to be the state-of-the-art in the ASR field~\cite{burchi2021, kim2022, peng2022, kim2023, rekesh2023}. We hypothesize that their ability to capture both temporal and spectral dependencies may serve as a good starting point for MEG-based speech tasks, which also exhibit structured spatiotemporal patterns over multiple timescales.

In this work, we evaluate a Conformer model adapted to the LibriBrain competition tasks. The models process raw MEG windows to predict either speech activity or phoneme identity, sharing the same backbone while differing in input and output processing. To ensure fair comparison and reproducibility, all experiments follow the official data splits and evaluation metrics provided by the organizers~\cite{landau2025}.

Our main contributions are fourfold:
(i) a unified Conformer architecture jointly optimized for both LibriBrain tasks;
(ii) a robust yet straightforward input normalization method that substantially improves holdout generalization;
(iii) an effective MEG-specific augmentation;
(iv) a dynamic grouping strategy to classify averaged samples in the Phoneme Task.

\ifinterspeechfinal
    This study demonstrates that adapting modern ASR architectures for MEG decoding produces competitive, robust results in both voice activity detection and phoneme recognition, emphasizing the growing convergence between speech processing and neural decoding research. For further implementation details, the technical documentation, source code, and checkpoints are available at \url{https://github.com/neural2speech/libribrain-experiments}.
\else
    This study demonstrates that adapting modern ASR architectures for MEG decoding produces competitive, robust results in both voice activity detection and phoneme recognition, emphasizing the growing convergence between speech processing and neural decoding research. For further implementation details, the technical documentation, source code, and checkpoints will be made publicly available; to preserve author anonymity, the link will be added after the review process.
\fi

\section{Related work}

Decoding speech from non-invasive neural signals has been explored with both Electroencephalography (EEG) and MEG using supervised learning, from linear baselines to deep architectures. EEG studies reconstruct speech features or spectrograms with CNN/Transformer-style models and subject-independent training, showing steady gains with more data~\cite{accou2023, xu2024}. On MEG, prior work demonstrated phrase and word-level decoding and trial-efficiency using wavelet denoising, CNNs, and transfer learning~\cite{dash2018_1, dash2018_2, dash2019}, as well as compact end-to-end networks adapted to sensor time series~\cite{sarma2024}, and transformer-based neural encoding that links linguistic context to MEG responses~\cite{klimovichgray2023}. Other studies have also started to explore phoneme-level decoding in perception and production modalities, comparing traditional and novel model architectures~\cite{dezuazo2024, dezuazo2025}. More recently, LibriBrain introduced an unprecedented amount of within-subject MEG data and standardized tasks for speech detection, phoneme classification, and keyword spotting~\cite{ozdogan2025, landau2025, elvers2025}. Beyond purely supervised setups, contrastive and self-supervised objectives, along with foundation-model guidance, have advanced non-invasive decoding across perception domains, improving retrieval and generation~\cite{defossez2023, benchetrit2024, banville2025}. Recent work on non-invasive brain-to-text systems combines discriminative decoders with language-model rescoring and cross-dataset scaling to produce significant score improvements~\cite{jayalath2025}.

\section{Dataset}\label{sec:dataset}

We evaluate our models on the LibriBrain dataset and PNPL 2025 competition setup~\cite{landau2025}. LibriBrain provides over 50 hours of within-subject, non-invasive MEG recorded from a single participant during naturalistic speech listening (Sherlock Holmes audiobooks), with time-aligned speech activity and phoneme annotations. Recordings were acquired with a 306-channel Elekta/MEGIN system (102 magnetometers, 204 planar gradiometers), originally sampled at \SI{1000}{Hz} and released with an official preprocessing pipeline (including bad-channel interpolation, head-position correction, signal-space separation, notch, and band-pass filtering), with model inputs downsampled to \SI{250}{Hz} following the competition protocol.

\begin{figure}
  \centering
  \includegraphics[width=\linewidth]{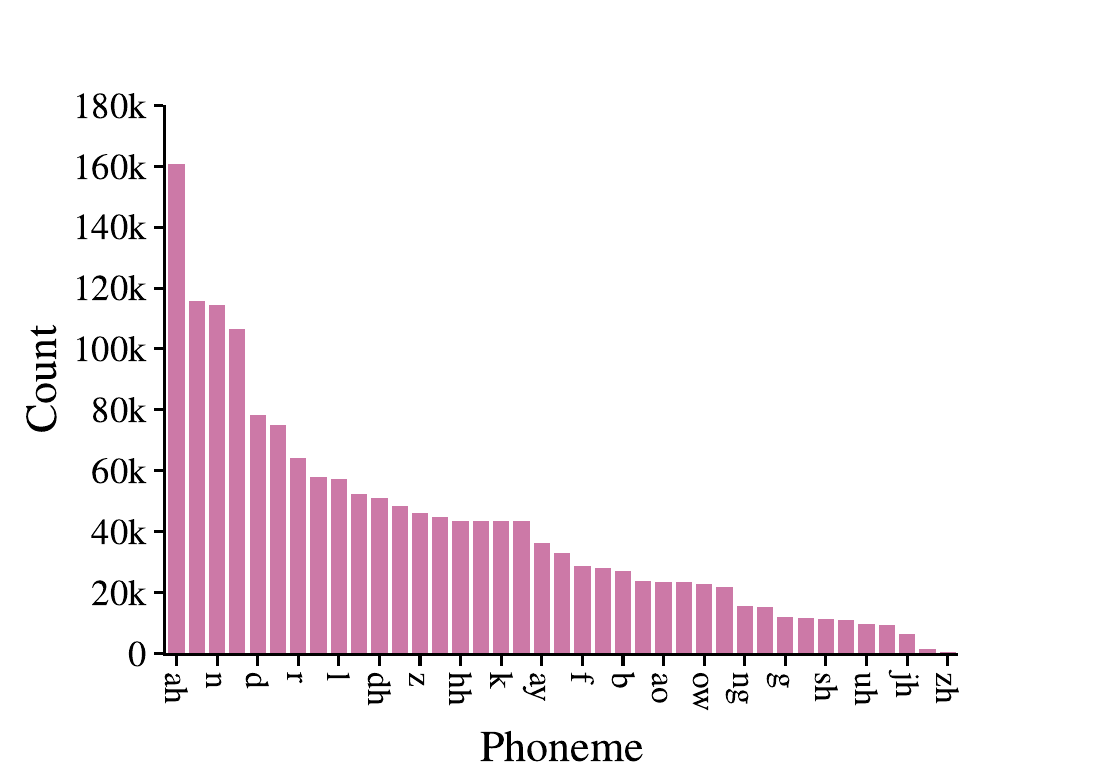}
  \caption{Total phoneme counts (before 100-sample averaging).}
  \label{fig:phoneme_counts}
\end{figure}

We follow the official Standard-track splits (train, validation, test, holdout) and evaluation metric (F1-macro) throughout, noting that the holdout split is not publicly available. For the Speech Detection task, labels are derived from voice-activity annotations over continuous recordings, resulting in a binary speech vs. silence classification problem. Across the full dataset, speech occupies \(\sim\)76.7\% of labeled time (40:19:38 speech vs.\ 12:16:37 silence). For the Phoneme Classification task, windows are aligned to 39 ARPAbet phoneme labels~\cite{arpabet}. Figure~\ref{fig:phoneme_counts} reports phoneme occurrence statistics for the phoneme windows, illustrating the long-tailed class distribution. Besides, in the competition setting for the phoneme task, holdout examples are provided as averages over 100 samples to improve SNR, and we mirror this setting in training via dynamic grouping (see Subsection~\ref{sec:phoneme}).

\section{Methods}

\subsection{Model}

We use a single Conformer encoder adapted to MEG for both tracks (Speech Detection and Phoneme Classification). The network takes raw sensor time series (306 channels) preprocessed with the official pipeline, where windowed signal segments are downsampled to \SI{250}{Hz}. For Speech, we use \SI{2.5}{s} windows (625 samples), and for the Phoneme Task, \SI{0.5}{s} windows (125 samples). A lightweight 1D convolutional projection adapts the 306 MEG channels to the Conformer input size (144), followed by a dropout of \(p{=}0.1\), a stack of Conformer blocks, and a linear classifier.

Each Conformer block follows the standard macaron layout: feed-forward, multi-head self-attention, depthwise temporal convolution, and a second feed-forward, with residual connections. We keep the design compact with a hidden size of 144. For Speech, based on Conformer Small, we use 16 layers, 4 heads, a hidden layer dimension of FFNs of 576, and the layer's depthwise convolution layer with a kernel size of 31. For Phonemes, we created a custom-sized Conformer with 7 layers, 12 heads, FFN dimension of 2048, and a kernel size of 127, to better adapt the model to the smaller dataset.

\subsection{Speech detection task specifics}\label{sec:speech}

\begin{figure*}
  \centering
  \includegraphics[width=\textwidth]{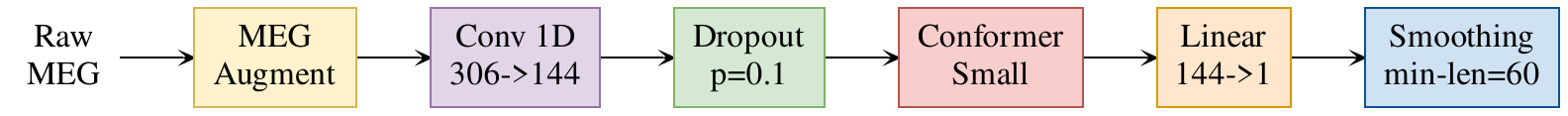}
  \caption{Conformer-based model architecture used for Speech Task.}
  \label{fig:model_speech}
\end{figure*}

The Speech model is a binary classifier with a single-logit head, binary cross-entropy loss~\cite{murphy2012}, and loss-level label smoothing of \(0.1\)~\cite{muller2019}. We found that simple, speech-style augmentation was sufficient for this task. Therefore, we develop and apply a lightweight MEG-specific variant of SpecAugment~\cite{park2019}, which we introduce as \textbf{MEGAugment}. It includes two operations: (i) \emph{time masking}, which zeroes two random temporal spans (max width \(T{=}180\) samples at \SI{250}{Hz}) per window; and (ii) \emph{bandstop masking}, which randomly suppresses narrow frequency bands (Theta, Alpha, Beta, Gamma, HGA~\cite{mandal2018}) using fourth-order infinite impulse response (IIR) notches with probability \(p{=}0.4\). Additionally, input windows slide with a 60-sample stride during training to increase diversity via overlapping.

The final layer outputs a probability of speech, which is smoothed during inference by removing speech sequences shorter than 60 samples (\SI{240}{ms}). The block diagram in Figure~\ref{fig:model_speech} summarizes the pipeline.

\subsection{Phoneme classification task specifics}\label{sec:phoneme}

\begin{figure*}
  \centering
  \includegraphics[width=\textwidth]{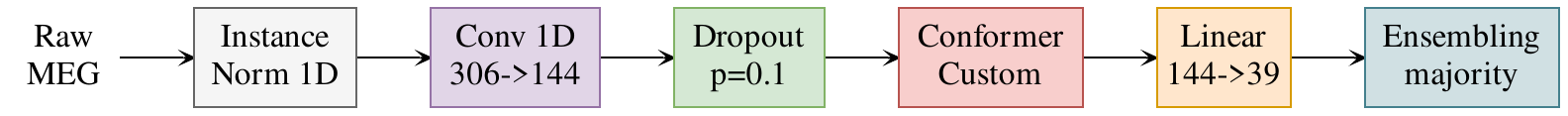}
  \caption{Conformer-based model architecture used for Phoneme Task.}
  \label{fig:model_phoneme}
\end{figure*}

The Phoneme model uses a linear classification head that outputs 39 logits and is trained with cross-entropy~\cite{bishop2006}. At inference, we select the class via \texttt{argmax}. To mitigate imbalance in the Phoneme Task, we use class weights \(w_c\) with the inverse square root of the number of samples (ISNS)~\cite{bakirarar2023} rule following Equation~\ref{eq:loss_weights_phoneme}, where \(n_c\) is the count of class \(c\) in the training set and \(C\) is the total number of classes.

\begin{equation} \label{eq:loss_weights_phoneme}
    w_c \propto \frac{1}{\sqrt{n_c}}, \qquad
    \frac{1}{C}\sum_c w_c = 1,
\end{equation}

The architecture (detailed in Figure~\ref{fig:model_phoneme}) includes an instance-level input normalization layer directly after the raw input~\cite{ulyanov2017, ulyanov2017_2}. For each averaged window \(x\in\mathbb{R}^{C\times T}\), it normalizes each channel independently using statistics computed over the time axis of that same window (i.e., no dataset or batch-level running mean/variance are accumulated), as defined in Equation~\ref{eq:instancenorm}.

\begin{equation}\label{eq:instancenorm}
y=\frac{x-\mathbb{E}[x]}{\sqrt{\mathrm{Var}[x]+\epsilon}}\cdot\gamma+\beta.
\end{equation}

Here, \(x\in\mathbb{R}^{C\times T}\) is one averaged MEG window (channels \(\times\) time), \(\mathbb{E}[x]\) and \(\mathrm{Var}[x]\) denote the mean and (biased) variance computed per window and per channel over the time axis (\(T\)), and \(\epsilon\) is a small constant for numerical stability. We use \texttt{InstanceNorm1d} without running statistics, and without affine parameters, so \(\gamma=1\) and \(\beta=0\).

This normalization is applied after the 100-sample averaging (as part of the model forward pass) and is used identically in training and evaluation. We find this choice essential for stable holdout performance: unlike validation and test, the holdout split exhibits different statistical characteristics, indicating distributional shift (see Subsection~\ref{sec:rms} for details). Instance normalization fundamentally removes amplitude and scale drift between windows and sessions. Consequently, it closes most of the holdout gap, even if it is sometimes slightly suboptimal on the other splits. Because the competition holdout recordings are pre-averaged over 100 samples to improve SNR, we adapt our training setup accordingly. Particularly, we use a 100-sample dynamic grouping loader that reshuffles groups each epoch, allowing the model to see many independent averages of the same class while preserving temporal locality.

Lastly, we ensemble the five best model seeds to smooth predictions and select the final phonemes using majority voting.

\section{Experimental setup}

We follow the official LibriBrain data partitions for all experiments, using the provided train, validation, test, and holdout splits. In our experience, re-partitioning or rebalancing the data did not produce consistent improvements, so we retain the original configuration throughout.

Training uses the AdamW optimizer~\cite{loshchilov2018} with a learning rate of \(1{\cdot}10^{-4}\), weight decay of \(5{\cdot}10^{-2}\), batch size of 256, and early stopping based on validation F1-macro with a patience of 10. We monitor F1-macro on the validation set to select the final checkpoint. These hyperparameters are shared across both tasks.

Evaluation follows the competition protocol, using F1-macro as the primary metric on test and holdout splits.

All model seeds are trained on individual NVIDIA A100 and H100 GPUs, and each model configuration is trained with ten random seeds for better comparison.

\section{Results}

Our Conformer-based models achieved competitive results in both tasks, as shown in Table~\ref{tab:leaderboard}. In Speech Detection, our best model reached an F1-macro of 88.9\%, surpassing the official baseline (68.0\%). In Phoneme Classification, our model reached 65.8\% F1-macro on holdout, which corresponds to the top result in the Standard track and the competition win\footnote{\url{https://neural-processing-lab.github.io/2025-libribrain-competition/prizes/}}. Notably, although we participated only in the Standard track, our phoneme results exceed those reported by systems submitted to the Extended track, which were allowed to use additional data sources. These results confirm that Conformer architectures, initially developed for ASR, transfer productively to MEG-based decoding and perform competitively with traditional convolutional and transformer models.

\begin{table*}
\centering
\caption{F1-macro scores in the holdout for both tasks (Standard track). Our model and scores are marked in bold.}
\label{tab:leaderboard}
\begin{tabular}{lrr}
  \toprule
  \multicolumn{1}{c}{\textbf{Model}} & \multicolumn{1}{c}{\textbf{Speech Detection \(\uparrow\)}} & \multicolumn{1}{c}{\textbf{Phoneme Classification \(\uparrow\)}} \\
  \midrule
      \textit{Naive Baseline} & \textit{45.30\%} & \textit{0.47\%} \\
      \textit{Baseline} & \textit{68.04\%} & \textit{6.10\%} \\
      10th place & 88.89\% & 60.96\% \\
      \textbf{MEGConformer} & \textbf{88.90\%} & \textbf{65.83\%} \\
      Best (Standard track) & 91.66\% & 65.83\% \\
  \bottomrule
\end{tabular}
\end{table*}

To contextualize these leaderboard results, we highlight two inference-time effects and the role of instance-level normalization. First, in the Speech task, the post-processing smoothing step produces a small but consistent improvement in F1-macro (from 88.12\% to 88.37\% in the test set), reaching a final holdout score of 88.90\% in Table~\ref{tab:leaderboard}. Second, in the Phoneme task, our single-model performance prior to ensembling is substantially lower (55.09\(\pm\)1.77\% across five seeds), and the final majority-vote ensemble is critical to achieving the reported 65.82\% on holdout, boosting scores by 19.48\%. Ultimately, instance-level normalization was essential for holdout generalization (over +200\%), outperforming batch (+17.8\%) and layer normalization (+88.2\%).

To better understand which components drive performance, and to characterize key dataset effects that impact generalization, we next report additional analyses: ablations with statistical testing, data-size scaling study, frequency band contribution analysis, a distribution-shift assessment across splits in the phoneme task, and experiments on phonetic-feature decoding.

\section{Additional analyses}\label{sec:additional_analyses}

\subsection{Ablation study and statistical significance}\label{sec:ablation_details}

For the ablation study, we start from our best-performing model and remove each improvement individually while keeping all other settings fixed. Each ablated variant is therefore identical to the full model except for one modification, allowing us to isolate the contribution of that component.

We report per-variant means and standard deviation over ten seeds and assess paired differences with the Wilcoxon signed-rank test~\cite{wilcoxon1945} over the F1-macro scores in the test set. This non-parametric test does not assume normality and evaluates whether the median of paired score differences differs from zero, indicating a genuine effect rather than random variation~\cite{santafe2015}. We use a p-value threshold of \(p \le 0.01\) to denote statistical significance.

\subsubsection{Speech task ablation}

\begin{table*}
\centering
\caption{Speech task variant test scores over ten seeds before smoothing. Bold marks the best scores.}
\label{tab:speech_ablation}
\begin{tabular}{lrrrrrr}
\toprule
\multicolumn{1}{c}{\textbf{Variant}} & \multicolumn{1}{c}{\textbf{F1\(_{\text{macro}}\) (\%)}} & \multicolumn{1}{c}{\textbf{F1\(_{\text{pos}}\) (\%)}} & \multicolumn{1}{c}{\textbf{Accuracy\(_\text{bal}\) (\%)}} & \multicolumn{1}{c}{\textbf{AUROC\(_\text{macro}\) (\%)}} & \multicolumn{1}{c}{\textbf{Jaccard (\%)}} \\
\midrule
\textbf{Our model}    & \textbf{87.59 \(\pm\) 0.70} & \textbf{94.72 \(\pm\) 0.23} & 85.72 \(\pm\) 1.19 & \textbf{96.15 \(\pm\) 0.32} & \textbf{78.64 \(\pm\) 1.02} \\
tmax=0.5     & 78.13 \(\pm\) 1.32 & 90.62 \(\pm\) 0.20 & 75.58 \(\pm\) 1.74 & 90.59 \(\pm\) 0.31 & 65.87 \(\pm\) 1.47 \\
SEANet       & 79.67 \(\pm\) 0.50 & 90.97 \(\pm\) 0.16 & 77.44 \(\pm\) 0.59 & 90.58 \(\pm\) 0.34 & 67.70 \(\pm\) 0.61 \\
No stride    & 85.18 \(\pm\) 0.55 & 93.06 \(\pm\) 0.14 & 83.59 \(\pm\) 1.12 & 94.35 \(\pm\) 0.20 & 75.01 \(\pm\) 0.75 \\
No augment   & 87.58 \(\pm\) 0.55 & 94.20 \(\pm\) 0.14 & \textbf{85.82 \(\pm\) 1.17} & 96.08 \(\pm\) 0.21 & 78.53 \(\pm\) 0.79 \\
\bottomrule
\end{tabular}
\end{table*}

As shown in Table~\ref{tab:speech_ablation}, window extension from 0.5 to \SI{2.5}{s} gives the largest gain (+10.8\% relative), and Conformer outperforms SEANet baseline~\cite{tagliasacchi2020} by +9.0\%. Both effects are significant: \(\text{tmax}=\SI{0.5}{s}\) vs.\ ours (\(W=0.0\), \(p=0.002\)), and SEANet vs. Conformer (\(W=0.0\), \(p=0.002\)). Reducing the training stride to 60 also significantly helps (+2.8\%; \(W=0.0\), \(p=0.002\)). Removing augmentation has a small effect (+0.01\%) and is not significant (\(W=13.0\), \(p=0.160\)). Although MEGAugment improvements are not significant in the final configuration, it did provide a significant gain of +1.8\% in earlier model versions, motivating its inclusion (\(W=1.0\), \(p=0.004\)).

\textbf{SEANet adjustments.}
To accommodate the longer input window \(\text{tmax}=\SI{2.5}{s}\) (625 samples at \SI{250}{Hz}), we modify only the temporal downsampler, the third \texttt{conv1d} (\(k{=}50\), \(s{=}25\) in the original), by setting its stride to \(s{=}160\). This keeps the intermediate length at \(L_\text{out}{=}4\), so the subsequent \(k{=}4\) stem still collapses to \(1\) (original: \(L_\text{out}=\lfloor(125-50)/25\rfloor+1=4\); adjusted: \(\lfloor(625-50)/160\rfloor+1=4\)). We also align the classifier and loss with the binary task by replacing the final \(1{\times}1\) \texttt{conv1d} from 512 to 2 with a single-logit 512 to 1 head and using BCE-with-logits plus label smoothing (\(0.1\)). Beyond architecture, we match the training and evaluation protocol used for our Conformer: sliding-window training with a stride of 60, MEGAugment, AdamW, early stopping on validation F1-macro, best-checkpoint selection, and validation and test scoring with a stride of 1.

\subsubsection{Phoneme task ablation}

\begin{table*}
\centering
\caption{Phoneme task test scores over ten seeds before ensembling. Bold marks the best scores.}
\label{tab:phoneme_ablation}
\begin{tabular}{lrrrrrr}
\toprule
\multicolumn{1}{c}{\textbf{Variant}} & \multicolumn{1}{c}{\textbf{F1\(_\text{macro}\) (\%)}} & \multicolumn{1}{c}{\textbf{Accuracy\(_\text{bal}\) (\%)}} & \multicolumn{1}{c}{\textbf{AUROC\(_\text{macro}\) (\%)}} \\
\midrule
\textbf{Our model}    & \textbf{44.29 \(\pm\) 2.84} & 47.75 \(\pm\) 2.70 & \textbf{96.67 \(\pm\) 0.31} \\
Fixed groups & 38.39 \(\pm\) 2.83 & 40.43 \(\pm\) 2.77 & 93.53 \(\pm\) 1.66 \\
No weights & 40.92 \(\pm\) 3.18 & 43.88 \(\pm\) 3.53 & 95.87 \(\pm\) 1.02 \\
Conformer Small & 44.09 \(\pm\) 4.56 & \textbf{48.49 \(\pm\) 4.19} & 96.33 \(\pm\) 0.60 \\
\bottomrule
\end{tabular}
\end{table*}

As shown in Table~\ref{tab:phoneme_ablation}, dynamic grouping (vs.\ fixed groups) gives the largest improvement (+13.3\% relative; \(W=1.0\), \(p=0.004\)). Inverse-square-root class weighting (our default) outperforms non-weighted loss by +7.6\% (not significant with \(W=10.0\), \(p=0.084\)). The custom Conformer provides a small lift over Conformer Small (+0.5\%; \(W=16.0\), \(p=0.844\)), consistent with similar capacity but with a better fit to the data characteristics.

\subsection{Data-size ablation}\label{sec:ablation}

\begin{figure}
  \centering
  \includegraphics[width=\linewidth]{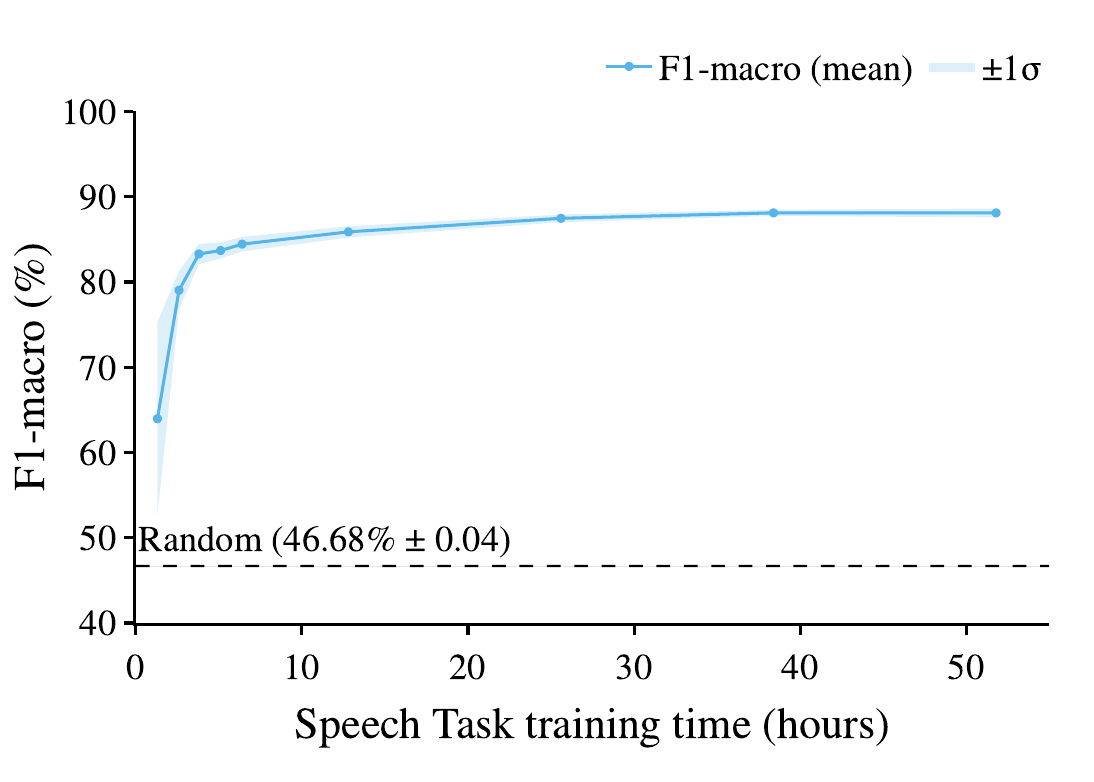}
  \caption{Data size ablation for Speech Detection Task.}
  \label{fig:ablation_speech}
\end{figure}

\begin{figure}
  \centering
  \includegraphics[width=\linewidth]{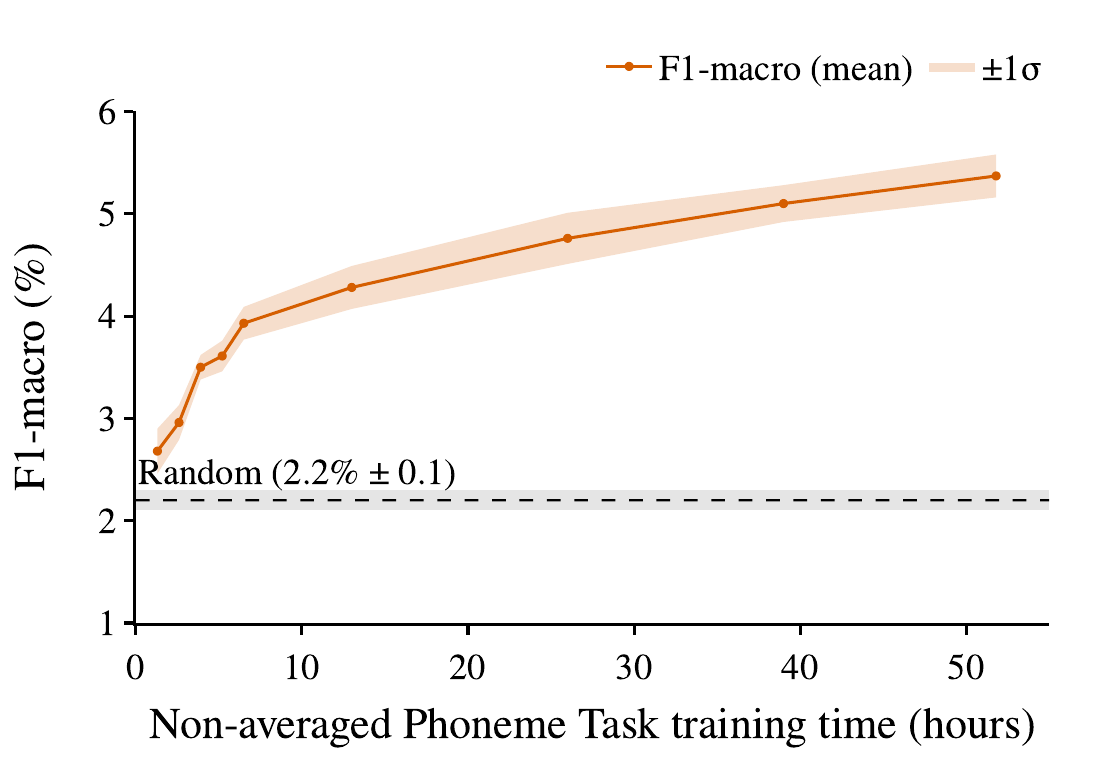}
  \caption{Data size ablation for Phoneme Classification Task, trained without sample averaging}.
  \label{fig:ablation_phoneme}
\end{figure}

To assess how the performance scales with within-subject MEG data, we conducted data-size ablations for both tasks using progressively larger subsets of the LibriBrain training set. Figures~\ref{fig:ablation_speech} and~\ref{fig:ablation_phoneme} show the resulting F1-macro scores as a function of the total hours of MEG data used for training (for phonemes, without sample averaging), with shaded bands indicating one standard deviation across random seeds.

For Speech Detection, performance rapidly improved with additional data and then began to saturate, suggesting that the task may already approach its ceiling with a single subject's data. In contrast, Phoneme Classification, even though with some diminishing returns signs~\cite{kaplan2020}, shows an upward trend with no apparent plateau, implying that larger-scale within-subject recordings could still give substantial gains. These results align with prior observations~\cite{ozdogan2025} that decoding performance continues to scale with the amount of high-quality MEG data available for training.

\subsection{Frequency band contribution to the tasks}\label{sec:frequency}

\begin{figure}
  \centering
  \includegraphics[width=\linewidth]{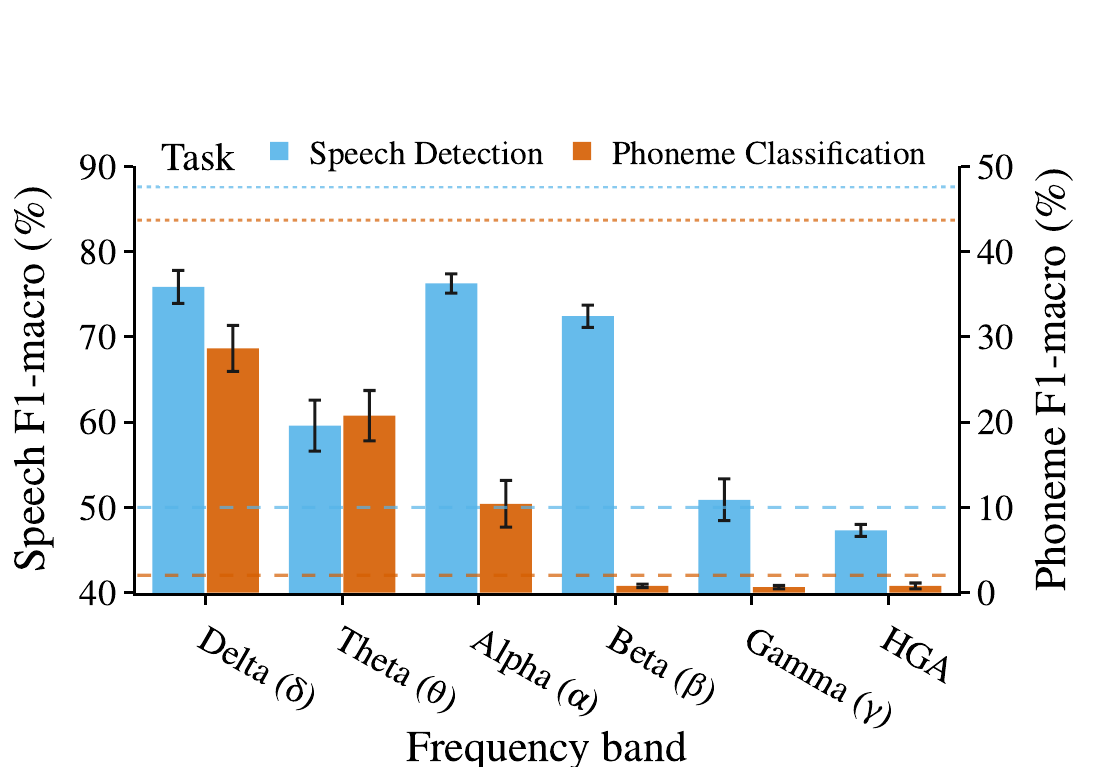}
  \caption{Decoding mean scores by frequency band for Speech and Phoneme Tasks (left/right y-axes), with black bars showing their standard deviation. The dashed lines represent the chance F1-macro level, and the dotted lines indicate the baseline (ceiling) F1-macro obtained without frequency filtering.}
  \label{fig:frequency}
\end{figure}

To probe which parts of the MEG spectrum are most informative for each task, we trained the same decoders while restricting the input to standard neurophysiological bands (Delta, Theta, Alpha, Beta, Gamma, and High-Gamma Activity; HGA)~\cite{mandal2018, gruenwald2023}. Figure~\ref{fig:frequency} summarizes the resulting F1-macro scores (mean\(\pm\)std over 10 seeds), together with chance and the unfiltered baseline (dotted line). We assess significance with one-sample Wilcoxon tests against chance.

In Speech Detection task, decoding is consistently above chance in low-frequency bands: Delta (75.86\%\(\,\pm\)1.94, \(W=55.0\), \(p<0.001\)), Theta (59.58\%\(\,\pm\)2.98, \(W=55.0\), \(p<0.001\)), Alpha (76.26\%\(\,\pm\)1.13, \(W=55.0\), \(p<0.001\)), and Beta (72.41\%\(\,\pm\)1.30, \(W=55.0\), \(p<0.001\)). In contrast, Gamma (50.88\%\(\,\pm\)2.45, \(W=38.0\), \(p=0.161\)) and HGA (47.29\%\(\,\pm\)0.70, \(W=0.0\), \(p=1.000\)) are at or below chance. Relative to the full-band baseline (87.59\%\(\,\pm\)0.70), the best single-band results (Alpha and Delta) retain \(\sim\)87\% of baseline performance, suggesting that, given the spatiotemporal sensitivity of MEG, speech/silence information accessible to non-invasive recordings is largely conveyed by slow cortical dynamics up to the Beta range, with higher frequencies being either less informative or less reliably measurable.

Similarly, in Phoneme Classification task, decoding is also significantly above chance in Delta (28.64\%\(\,\pm\)2.70, \(W=55.0\), \(p<0.001\)), Theta (20.75\%\(\,\pm\)2.95, \(W=55.0\), \(p<0.001\)), and Alpha (10.41\%\(\,\pm\)2.74, \(W=55.0\), \(p<0.001\)), but falls below chance in Beta, Gamma and HGA, all being not significant (\(W=0.0\), \(p=1.000\)). All band-limited models remain substantially below the unfiltered baseline (43.69\%\(\,\pm\)4.76): Delta retains \(\sim\)66\% of baseline, Theta \(\sim\)48\%, and higher bands contribute little in isolation. Significance tests confirm that each band-restricted setting is worse than the baseline in both tasks (\(W=0.0\), \(p=1.000\)), indicating that speech decoding benefits from combining information across frequency ranges rather than relying on a single band. However, this effect should be interpreted with caution, as the averaging applied in the phoneme task might have suppressed higher-frequency components.

Overall, both tasks emphasize low-frequency components, with negligible standalone contribution from Gamma and HGA in this dataset. This pattern is consistent with MEG evidence that low-frequency (Delta/Theta) activity supports speech decoding~\cite{cogan2011}, whereas higher-frequency information appears more accessible in invasive recordings~\cite{proix2022}, and also aligns with earlier frequency-based MEG decoding analyses~\cite{dezuazo2025}.

\subsection{Distribution shift across splits in the phoneme task}\label{sec:rms}

To explain the holdout generalization gap observed in the phoneme task, we quantify scale mismatches between partitions by computing the per-window root-mean-square (RMS) energy across channels and time as in Equation~\ref{eq:rms}. The \(x\) is the MEG window (\(C{=}306\) channels, \(T{=}125\) samples). For validation and test, windows were read with the standard competition preprocessing and standardization enabled. The holdout was read as stored (already standardized on disk). We then formed outline-only histograms using the same bin edges and density normalization across splits.

\begin{equation} \label{eq:rms}
  \mathrm{RMS}(x) = \sqrt{{mean}\big(x^2\big)} \quad\text{for } x\in\mathbb{R}^{C \times T},
\end{equation}

\begin{figure}
  \centering
  \includegraphics[width=0.9\linewidth]{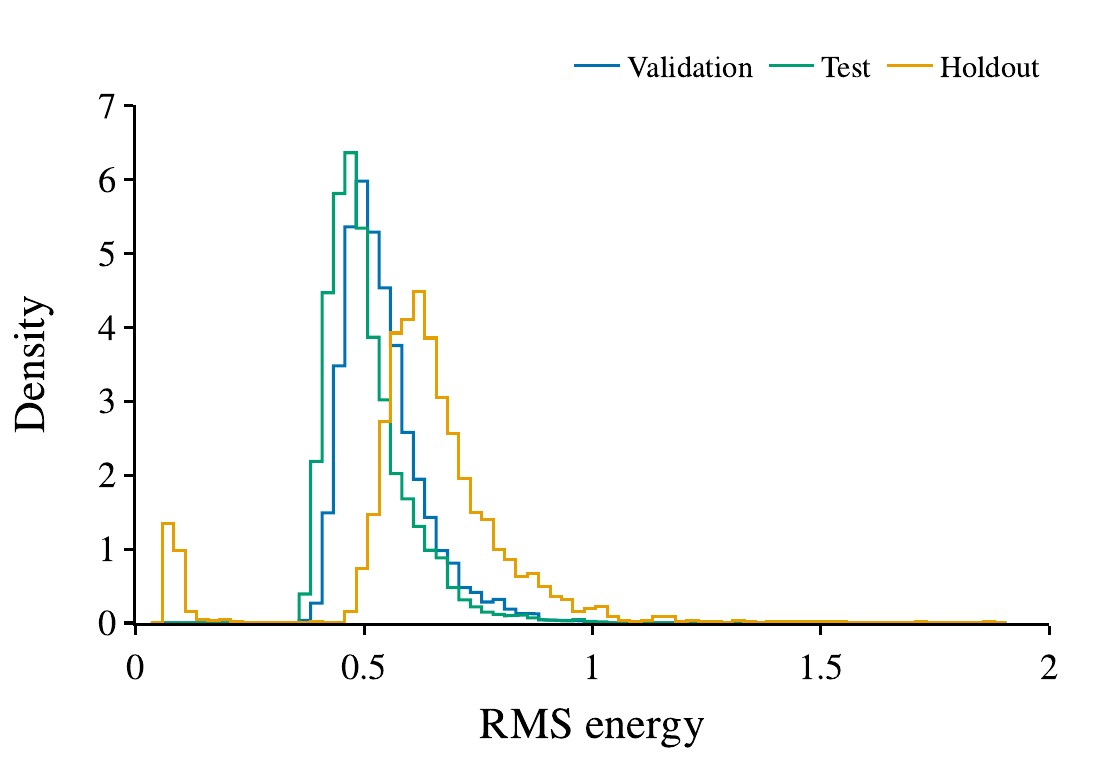}
  \caption{RMS energy distribution per split in Phoneme Task.}
  \label{fig:rms_histogram}
\end{figure}

Figure~\ref{fig:rms_histogram} shows that the holdout distribution is bimodal (two peaks near \(0.08\) and \(0.62\)), with markedly larger dispersion (mean \(0.64\), standard deviation \(0.22\)), whereas validation and test are unimodal and tighter (validation \(0.54 \pm 0.09\), test \(0.51 \pm 0.09\)). This shift motivated the instance-level normalization used in the Phoneme model to remove amplitude and scale differences window-by-window, thereby improving robustness on the holdout.

\subsection{Phonetic-feature decoding as a linguistically grounded probe}\label{sec:features}

\begin{table*}
\centering
\caption{Decoding performance for binary phonetic features in the non-averaged test set (in \%).}
\label{tab:feature_results}
\begin{tabular}{lrrrrr}
\toprule
\multicolumn{1}{c}{\textbf{Feature}} & \multicolumn{1}{c}{\textbf{Positives (\%)}} & \multicolumn{1}{c}{\textbf{F1\(_{\text{macro}}\) (\%)}} & \multicolumn{1}{c}{\textbf{F1\(_{\text{pos}}\) (\%)}} & \multicolumn{1}{c}{\textbf{Accuracy\(_\text{bal}\) (\%)}} & \multicolumn{1}{c}{\textbf{AUROC\(_\text{macro}\) (\%)}} \\
\midrule
\textit{Speech} & \textit{76.76} & \textit{88.10 \(\pm\) 0.52} & \textit{94.42 \(\pm\) 0.17} & \textit{86.38 \(\pm\) 0.97} & \textit{96.38 \(\pm\) 0.18} \\
\textit{Phoneme} & \textit{--} & \textit{5.37 \(\pm\) 0.21} & \textit{4.00 \(\pm\) 1.08} & \textit{5.86 \(\pm\) 0.22} & \textit{64.35 \(\pm\) 1.05} \\
\midrule
Voicing & 77.38 & 57.77 \(\pm\) 0.56 & 83.23 \(\pm\) 1.30 & 57.38 \(\pm\) 0.69 & 63.88 \(\pm\) 0.45 \\
Plosive & 18.82 & 55.58 \(\pm\) 0.53 & 25.54 \(\pm\) 1.86 & 55.31 \(\pm\) 0.57 & 62.09 \(\pm\) 1.22 \\
Fricative & 18.55 & 53.78 \(\pm\) 0.78 & 23.85 \(\pm\) 2.07 & 53.70 \(\pm\) 0.71 & 57.71 \(\pm\) 1.34 \\
Affricate & \textcolor{red}{0.97} & \textcolor{red}{49.69 \(\pm\) 0.03} & \textcolor{red}{0.00 \(\pm\) 0.00} & \textcolor{red}{49.98 \(\pm\) 0.06} & \textcolor{red}{60.29 \(\pm\) 1.92} \\
Nasal & 11.26 & 51.74 \(\pm\) 0.54 & 12.98 \(\pm\) 1.06 & 51.67 \(\pm\) 0.50 & 55.97 \(\pm\) 1.47 \\
Liquid & 7.51 & 50.80 \(\pm\) 1.02 & 8.16 \(\pm\) 3.03 & 51.02 \(\pm\) 0.57 & 55.81 \(\pm\) 1.20 \\
Glide & 3.62 & 52.16 \(\pm\) 0.81 & 6.70 \(\pm\) 1.65 & 51.83 \(\pm\) 0.70 & 60.93 \(\pm\) 2.25 \\
\bottomrule
\end{tabular}
\end{table*}

As an additional analysis to probe linguistically grounded representations for MEG decoding, we trained separate binary classifiers targeting core phonetic features related to manner of articulation and voicing without sample averaging. Each model was adapted from the Phoneme Classification configuration described in Subsection~\ref{sec:phoneme}, using the same backbone and training hyperparameters, but modified for binary output. Notably, we replaced the 39-class head with a single-logit output, used a binary cross-entropy loss with label smoothing and automatic positive-class weighting following Equation~\ref{eq:loss_weights_phoneme} to address class imbalance (based on the Speech model in Subsection~\ref{sec:speech}).

This approach follows the ideas proposed by the competition authors in their blog on language-inspired phoneme classification~\cite{kwon2025}, where phonetic features are suggested to handle dataset imbalance and improve infrequent phoneme decoding.

Based on~\cite{mesgarani2014}, we defined six phonetic features based on manner of articulation: Plosive, Fricative, Affricate, Nasal, Liquid, Glide, and Voicing that distinguishes voiced from voiceless phonemes, including vowels and consonants. Table~\ref{tab:feature_results} reports preliminary decoding results using F1-macro, positive-class F1, balanced accuracy, and AUROC (mean~\(\pm\)~std across ten seeds).

Following Subsection~\ref{sec:ablation_details}, statistical significance is evaluated with a one-sample Wilcoxon signed-rank test against chance (\(0.5\) for F1-macro).

Among these features, Voicing showed the strongest decoding signal, approaching 58\% F1-macro (significant with \(W=55.0\), \(p=0.001\)). Despite its similarity in label balance to the Speech Task, scores are not on par, implying that decoding some fine-grained speech features may not be straightforward. Manner-based features such as Plosive and Fricative were moderately decodable (both significant with \(W=55.0\), \(p=0.001\)), while Affricate, a composite and under-represented category (0.97\%), remained at chance (not significant with \(W=0.0\), \(p=1.0\)). Attempts to address the affricate issue through ensembling of Plosive and Fricative models, transfer learning from these features, and larger effective batch sizes (via gradient accumulation) did not improve the convergence. This supports the view that decomposing phonemes into binary linguistic features alone is insufficient to overcome the data scarcity, as some linguistic features may also be rare enough not to be easily decodable, like affricates here.

Overall, these experiments suggest that feature-based representations are promising for improving interpretability and possibly scaling, but further research is needed to address under-represented classes and explore multitask formulations that jointly learn shared articulatory subspaces.

\section{Discussion and conclusion}

The Conformer architecture proves suitable to decode the spatiotemporal nature of MEG signals, combining convolutional blocks that capture local temporal structure with self-attention for long-range dependencies. Using very similar models across both tasks, we achieved top-10 leaderboard performance, with minor input preprocessing differences between Speech and Phoneme decoding. In particular, our approach won the Phoneme Classification track in the LibriBrain Competition 2025. Instance-level normalization is indispensable to mitigate distributional shifts between training and holdout splits in the Phoneme Task, substantially improving generalization. The additional analyses in Section~\ref{sec:additional_analyses} suggest that this robustness is closely tied to distribution shift across splits. Moreover, our frequency-band analysis shows that both speech and phoneme decoding draw most of their discriminative power from low-frequency components (Delta--Beta), while Gamma and HGA contribute little in isolation. This suggests that, in the LibriBrain setting, effective MEG decoding benefits from integrating information across slow cortical dynamics, and high-frequency activity contribution is insignificant. Furthermore, a simple SpecAugment-based~\cite{park2019} augmentation was effective for speech detection, but showed limited benefit for the more complex, imbalanced phoneme classification task. On the other hand, as LibriBrain is single-subject, cross-subject generalization remains an open challenge for future work.

Looking ahead, adapting speech-model architectures for MEG decoding could open the way toward end-to-end speech reconstruction using sequential objectives such as CTC or seq2seq heads, and even enable speech synthesis. Another promising direction is the use of linguistically grounded feature-based representations~\cite{kwon2025}, which may help tackle data imbalance and improve interpretability by decomposing phoneme classification into binary articulatory features (see Subsection~\ref{sec:features} for the challenges found). Finally, our scaling experiments (Subsection~\ref{sec:ablation}) indicate that while speech decoding performance starts to saturate, phoneme classification continues to improve with additional data, suggesting further gains could be achieved as larger MEG datasets become available.

\ifinterspeechfinal
    \section{Acknowledgements}

    The Basque and Spanish Governments fund this research (IKUR-IKA-23/18, and AIA2025-163317-C31 by MICIU/AEI / 10.13039/501100011033/). The authors thank both the DIPC Supercomputing Center and the technological management body of the Basque Government, EJIE, for their technical and human support. This work was conducted as part of the \#neural2speech and brAIn2lang teams. Last but not least, special thanks to Ekain Arrieta for inspiring the application of Conformer and ASR-based architectures to MEG decoding.
\fi

\bibliographystyle{IEEEtran}
\bibliography{BibEntries}

\end{document}